\documentclass[letterpaper,10pt,journal,twoside]{IEEEtran}
\def\FinalRoot{}

\IEEEoverridecommandlockouts                              

\newif\ifanon\anonfalse
\newif\ifarxiv
\ifdefined\buildarxiv
  \arxivtrue
\else
  \arxivfalse
\fi
\arxivtrue
\newif\ifshowappendix
\showappendixfalse
\ifanon\showappendixfalse\fi
\ifarxiv\showappendixfalse\fi
\newif\ifdiffmarks
\diffmarkstrue
\ifdiffmarks\else\AtBeginDocument{\colorlet{blue}{black}}\fi
\usepackage{graphicx}
\usepackage{amsmath, amssymb}
\usepackage{cite}
\usepackage{algorithm, algorithmic}
\usepackage{booktabs}
\usepackage{multirow}
\usepackage[hidelinks]{hyperref}
\usepackage[capitalize, noabbrev]{cleveref}
\crefname{figure}{Fig.}{Figs.}
\Crefname{figure}{Fig.}{Figs.}
\crefname{section}{Section}{Sections}
\Crefname{section}{Section}{Sections}
\crefname{table}{Table}{Tables}
\Crefname{table}{Table}{Tables}
\crefname{equation}{Eq.}{Eqs.}
\Crefname{equation}{Eq.}{Eqs.}

\hypersetup{pdftitle={Flow-ERD: Agent-type Aware Flow Matching With Entropy-Regularized Distillation for Diverse Traffic Simulation},pdfauthor={Seulbin Hwang, Kiyoung Om, Dae Jung Kim, Jinhan Lee}}
\ifanon\hypersetup{pdfauthor={}}\fi

\usepackage{xcolor}
\colorlet{blue}{black}
\definecolor{FinalBlue}{rgb}{0,0.15,0.65}
\newif\ifFinalMarks
\ifdefined\FinalMarkChanges\FinalMarkstrue\else\FinalMarksfalse\fi
\newcommand{\finalchange}[1]{\ifFinalMarks{\color{FinalBlue}#1}\else#1\fi}
\title{
Flow-ERD: Agent-type Aware Flow Matching \finalchange{With} Entropy-Regularized Distillation for Diverse Traffic Simulation
}

\ifanon
\author{Anonymous Author(s)%
\thanks{Author names and affiliations omitted for double-anonymous review.}%
}
\else
\author{
    Seulbin~Hwang\smash{\textsuperscript{*}},~
    Kiyoung~Om\smash{\textsuperscript{*}},~
    Dae~Jung~Kim,~
    and~Jinhan~Lee\smash{\textsuperscript{\dag}}%
\thanks{Manuscript received: July 3, 2026; Revised: September 6, 2026; Accepted: October 13, 2026. This paper was recommended for publication by Editor Sehoon Ha upon evaluation of the Associate Editor and Reviewers' comments. \finalchange{This research was supported by a grant (code RS-2023-00233952) from the R\&D Program funded by the Ministry of Land, Infrastructure and Transport of the Korean government.}}%
\thanks{All authors are with NAVER LABS \finalchange{Corporation}, \finalchange{Seongnam-si, Gyeonggi-do 13561, }Republic of Korea
        (e-mail: \{h.sb, se99an, daejung.kim, \mbox{jinhan.lee}\}@naverlabs.com).}%
\thanks{\textit{\smash{\textsuperscript{*}}Seulbin Hwang and Kiyoung Om are co-first authors and contributed equally to this work.}}%
\thanks{\textit{\smash{\textsuperscript{\dag}}Corresponding author: Jinhan Lee.}\newline
(E-mail: jinhan.lee@naverlabs.com).}%
}
\fi

\AddToHook{shipout/foreground}{%
  \ifnum\value{page}=1\relax
    \put(0,-175){\makebox[\paperwidth][c]{\normalfont\fontsize{9}{11}\selectfont
      \textbf{Project page \& code: }\href{https://seulbinhwang.github.io/flow-erd-project-page/}{\textcolor{FinalBlue}{\nolinkurl{https://seulbinhwang.github.io/flow-erd-project-page/}}}}}%
    \put(49.147,-761){\makebox(0,0)[lt]{\parbox[t]{516.0pt}{\normalfont\fontsize{7}{8}\selectfont
      \textcopyright~2026 IEEE. Personal use of this material is permitted. Permission from IEEE must be obtained for all other uses, in any current or future media, including reprinting/republishing this material for advertising or promotional purposes, creating new collective works, for resale or redistribution to servers or lists, or reuse of any copyrighted component of this work in other works.}}}%
  \fi
}

\begin{document}

\maketitle

\begin{abstract}
Realistic and diverse traffic simulation is essential to autonomous driving development. Yet prevailing benchmarks predominantly reward realism, and recent methods have optimized accordingly, leaving diversity underexplored. We introduce \textbf{Flow-ERD}, a multi-agent simulator that pursues realism and diversity jointly. Its backbone, \textbf{Agent-Type Aware Flow Matching} (AFM), couples flow matching's multi-modal expressiveness with type-specific kinematic execution. It preserves fine-grained diversity while keeping motions consistent with each agent type. A second stage, \textbf{Entropy-Regularized Distillation} (ERD), fine-tunes the closed-loop rollout distribution with an entropy-regularized reverse-KL objective. This mitigates \finalchange{closed-loop} covariate shift while explicitly preventing collapse onto high-density modes. We evaluate Flow-ERD with \finalchange{Cross-Pair Diversity and its validity-gated variant} alongside standard realism scores. \finalchange{Validation experiments further show that tuning entropy regularization recovers most of the pretrained backbone's diversity while retaining realism gains.} Flow-ERD ranks first \finalchange{in test-split realism} on the \finalchange{Waymo Open Sim Agents Challenge (WOSAC)} and dominates the \finalchange{validation} realism--diversity Pareto front among reproducible baselines.
\end{abstract}


\section{Introduction}
\label{sec:intro}

\begin{figure}[!t]
  \centering
  \includegraphics[width=0.70\linewidth,trim=0pt 0pt 0pt 12pt,clip]{\FinalRoot 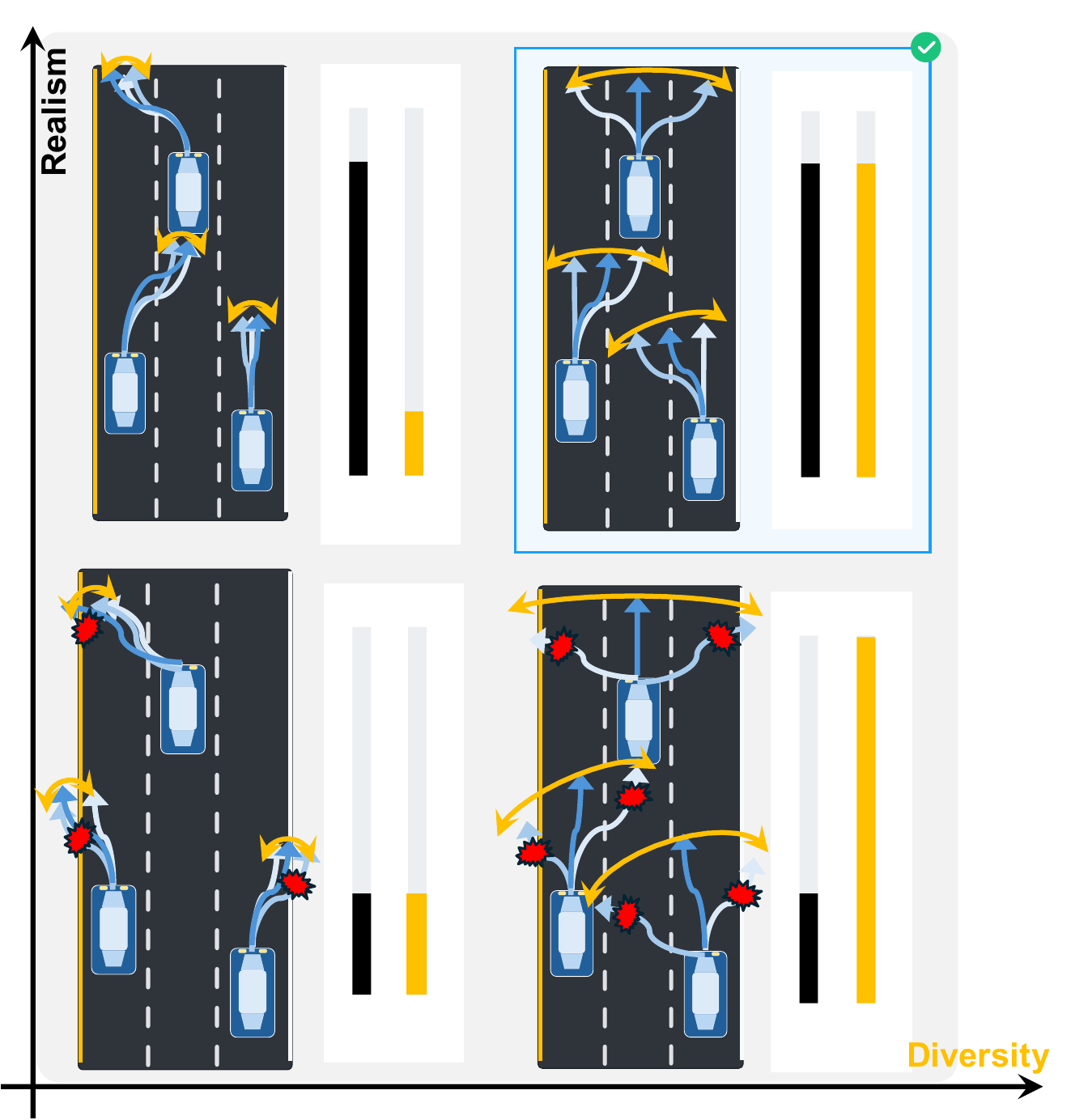}
  \caption{Low-diversity rollouts concentrate on a dominant behavior, whereas low-realism rollouts deviate from plausible traffic motion. Flow-ERD targets the desired regime of realistic and diverse closed-loop rollouts.}
  \label{fig:teaser}
  \vspace{-20pt}
\end{figure}

\IEEEPARstart{T}{raffic} simulation has become core infrastructure for autonomous driving, supporting controlled validation before public-road deployment as well as the development of AV planning policies~\cite{dosovitskiy2017carla, gulino2023waymax}. For simulation to serve these roles, the surrounding agents, including vehicles, cyclists, and pedestrians, must be \emph{realistic}, imitating real-world traffic behavior and reacting to one another in a closed loop; and \emph{diverse}, spanning the multiple plausible futures of a scene to ensure the ego policy's robustness~\cite{montali2023waymo, suo2021trafficsim}. These properties must hold jointly, not as alternatives (\cref{fig:teaser}).

Generative models are well-suited to capturing these properties, learning the distribution of traffic-agent behavior from large-scale data~\cite{Ettinger_2021_ICCV} with expressive architectures~\cite{radford2018improving, ho2020denoising}. Indeed, recent learning-based simulators~\cite{philion2024trajeglish, wu2024smart, zhang2026trajtok, huang2026versatile, jiang2024scenediffuser, huang2025mdg, lin2025revisitmixturemodelsmultiagent} have made substantial progress on realism, especially under benchmarks such as the Waymo Open Sim Agents Challenge (WOSAC)~\cite{montali2023waymo}. The benchmark's realism score, however, is measured against a single logged future, and thus cannot distinguish a model that merely fits the logged future from one that captures diverse plausible behaviors.
As models are increasingly optimized for this benchmark, diversity has been acknowledged but rarely treated as equally important as realism: it is often left to a sampling hyperparameter~\cite{wu2024smart} or assessed only qualitatively~\cite{zhang2025closed, pei2026advancing}.

This gap also manifests in how simulators are designed and trained. By design, backbones trade realism against diversity. Next-token-prediction based methods~\cite{philion2024trajeglish, wu2024smart, zhang2026trajtok, zhao2024kigras} draw each action from a predefined discrete vocabulary derived from logged data. Because they encode patterns already present in the data, they provide an inductive bias toward realistic, type-compatible motion. However, the fixed vocabulary collapses fine-grained motion onto a coarse set of tokens, which inherently bounds the attainable diversity~\cite{zhang2026trajtok}. Continuous representations~\cite{Zhong2023CTG, huang2026versatile, jiang2024scenediffuser}, especially diffusion models, remove this bottleneck and are effective at modeling complex multi-modal distributions~\cite{liao2025diffusiondrivetruncateddiffusionmodel}, yet this expressiveness does not guarantee realism. Continuous policies must learn valid motion implicitly, so occasional type-incompatible predictions can be fed back into the closed-loop context and amplify rollout errors.

On the training side, backbones trained in open loop are deployed in closed loop, leading to covariate shift~\cite{ross2011reduction}, in which a model's own actions feed into the next prediction and accumulate error over horizons. Existing methods address this by augmenting the data with log-proximal rollouts~\cite{zhang2025closed, garcia2025road, chang2025langtraj}, by using reinforcement learning to maximize a realism score~\cite{ahmadi2026rlftsim, pei2026advancing}, or by using a generative adversarial reward~\cite{guo2026decompgail}. While these methods reduce compounding closed-loop error, it remains unclear whether they do so by narrowing the distribution of generated behaviors.

To this end, we introduce \textbf{Flow-ERD}, a multi-agent simulator that jointly addresses realism and diversity. It is built on an \textbf{Agent-type Aware Flow-Matching} (AFM) backbone which adopts flow matching~\cite{lipman2022flow}---a diffusion-family continuous generator---for its multi-modal expressiveness, and grounds it in kinematics to keep that diversity realizable. During closed-loop rollout, AFM samples continuous actions and executes them through agent-type-specific transitions, avoiding token-codebook limits and type-incompatible unconstrained motion.

The second component, \textbf{Entropy-Regularized Distillation} (ERD), preserves the diversity while addressing the covariate shift problem. ERD fine-tunes the closed-loop rollout distribution with an entropy-regularized reverse-KL objective that drives it toward the data distribution while mitigating collapse toward high-density modes.

We evaluate Flow-ERD on the WOSAC using the standard realism meta metric (RMM) together with our proposed \textbf{Cross-Pair Diversity} (CPD), a log-free metric that measures rollout spread across multiple samples. This joint evaluation establishes a comprehensive realism--diversity landscape, setting the stage for our core findings.

Our contributions are threefold:
\begin{itemize}
    \item We propose Agent-Type Aware Flow Matching
    (AFM), which captures fine-grained multi-modal behaviors through continuous flow matching while preserving realism through agent-type aware transitions.
    \item We introduce Entropy-Regularized Distillation (ERD), a closed-loop fine-tuning method that mitigates covariate shift while explicitly preserving multi-modality.
    \item On the WOSAC~\cite{montali2023waymo} test benchmark, AFM achieves the state-of-the-art kinematic score, and Flow-ERD ranks first overall in realism; on the validation split, both attain the highest rollout diversity among reproducible baselines, dominating the realism--diversity Pareto front.
\end{itemize}

\section{Related Works}
\label{sec:related}
\subsection{Learning-Based Multi-Agent Simulation}
Next-token prediction (NTP)- based models have been among the strongest WOSAC submissions~\cite{philion2024trajeglish, wu2024smart, zhou2024behaviorgpt, zhang2026trajtok, zhao2024kigras}. Their discrete vocabularies provide an inductive bias toward realism by composing motion from data-derived or rule-based primitives that preserve type-specific motion patterns for pedestrians, vehicles, and cyclists. This constrains agents to data-supported, type-compatible motions and prohibits feeding implausible predictions back into their input conditions. However, this also limits diversity. Any motion absent from the vocabulary can only be approximated by its nearest available token, bounding the generated behaviors by the coverage of the token vocabulary~\cite{zhang2026trajtok}. 

A natural remedy is to replace discrete tokens with continuous outputs. UniMM~\cite{lin2025revisitmixturemodelsmultiagent} unifies discrete NTP-based models and continuous mixture models under a common mixture-model view, covering both anchor-free and anchor-based variants. However, continuity alone is not sufficient for diversity, as continuous mixture models still capture multimodality through a finite set of components, and anchor-based variants further tie coverage to predefined anchors, limiting their expressiveness.

Diffusion-based simulators model joint futures in continuous space, naturally supporting multimodality and controllability without committing to a fixed number of modes or anchors~\cite{huang2026versatile, jiang2024scenediffuser, Zhong2023CTG, tan2025prosim, chang2025langtraj, zhou2025nexus}. Although they remove the fixed-vocabulary bottleneck, models must learn valid motion supports implicitly from data. Naive modeling of its continuous state or action representation may therefore generate type-incompatible motions, such as lateral slip for vehicles and cyclists or unnecessarily constrained pedestrian behavior. Fed back as closed-loop history, such states may amplify distribution shift more readily than in token-based systems. 
Our method samples multi-modal continuous kinematic actions and executes them through type-specific transitions, yielding type-compatible motion.

\subsection{Mitigating Covariate Shift in Traffic Simulation}
Learning-based traffic simulators are typically pretrained by behavior cloning (BC), conditioned on logged histories during training but on their own generated states at deployment, causing the standard covariate-shift problem~\cite{ross2011reduction,ho2016generative}. Recent methods therefore fine-tune pretrained models on closed-loop rollouts to improve realism. One line constructs supervision from log-proximal rollouts: CAT-K~\cite{zhang2025closed} selects the top-$K$ action token whose next state is closest to the ground truth (GT) and trains on the resulting DaD-style recovery tokens~\cite{venkatraman2015improving}; RoaD~\cite{garcia2025road} samples many rollouts, filters them by GT distance, and treats the retained rollouts as demonstrations; and LangTraj~\cite{chang2025langtraj} adapts this principle to diffusion by denoising candidates from lightly noised GT trajectories and learning recovery actions.

Another line directly optimizes realism-oriented objectives. SMART-R1~\cite{pei2026advancing} and RLFTSim~\cite{ahmadi2026rlftsim} apply reinforcement fine-tuning on the WOSAC Realism Meta Metric (RMM) and DecompGAIL~\cite{guo2026decompgail} stabilizes GAIL~\cite{ho2016generative} in multi-agent settings by decomposing the discriminator into ego--map and ego--neighbor terms.

Although these methods obtain supervision differently, they primarily pull rollouts toward the recorded trajectory, RMM statistics, or the logged-data manifold. This improves closed-loop stability, but leaves scenario-conditioned diversity implicit. In contrast, our method fine-tunes on the model's own rollouts with an entropy-regularized distribution-matching objective that balances covariate-shift reduction with multimodality preservation.

\section{Preliminaries}
\label{sec:prelim}
\subsection{Multi-Agent Driving Simulation}
\label{subsec:problem_setup}
A scenario is specified by a road map $\mathcal M$ and $N$ traffic participants indexed by $i\in\{1,\ldots,N\}$, each with agent type $c_i\in\mathcal C=\{\textsc{veh},\textsc{cyc},\textsc{ped}\}$. Agent $i$ at time $t$ has modeled planar state $\mathbf s_t^i=(\mathbf p_t^i,\psi_t^i)\in\mathcal{S}$, where $\mathbf p_t^i\in\mathbb R^2$ is the agent's center position, $\psi_t^i$ is the heading and $\mathbf b^i=(\ell^i,w^i)$ denotes the 2D box size. 
The joint scene state is $\mathbf s_t=(\mathbf s_t^1,\ldots,\mathbf s_t^N)$. We denote the historical scene context by $\mathcal H_t=(\mathbf s_{t-L+1:t},\mathcal M)$,
 where $L$ is the history length, and let $\mathcal H_0$ be the initial history context. 
Starting from $\mathcal H_0$, a simulator evolves the scene over horizon $T$ and produces a state rollout $\tau=(\mathbf s_1,\ldots,\mathbf s_T)$. Simulation is closed-loop: each realized scene becomes the context for the next prediction.

\subsection{Holonomic and Bicycle-Style Motion}
\label{subsec:transition}
Traffic-agent motion models commonly distinguish holonomic
planar motion for freely moving participants, such as pedestrians,
from bicycle-style non-holonomic motion for wheeled road users~\cite{paden2016survey,polack2018}.
The former permits both longitudinal and lateral
displacement in the agent frame, whereas the latter does not treat
lateral motion as an independent executed input. 

For heading
$\psi$, let $R(\psi)$ be the planar rotation matrix,
$\mathbf u_\parallel(\psi)=(\cos\psi,\sin\psi)^\top$,
$\mathbf u_\perp(\psi)=(-\sin\psi,\cos\psi)^\top$, and
$\operatorname{sinc}(z)=\sin(z)/z$ with $\operatorname{sinc}(0)=1$.
We denote $a_\parallel$, $a_\perp$, and $a_\psi$ for one-step
longitudinal displacement, lateral displacement, and heading
change in the agent frame.

A holonomic displacement executes both displacement channels:
\begin{equation}
\Delta\mathbf p_{\rm hol}
=
R(\psi)
\begin{bmatrix}
a_\parallel\\
a_\perp
\end{bmatrix}.
\end{equation}

For bicycle-style motion, let $r$ be the no-slip offset, the scalar
distance from the box center to the no-slip reference point where
lateral motion is suppressed. With midpoint heading
$\bar{\psi}=\psi+a_\psi/2$, the box-center displacement is
\begin{equation}
\begin{aligned}
\Delta\mathbf p_{\rm nh}
&= \Delta\mathbf p_{\rm prog}+\Delta\mathbf p_{\rm swing},\\
\Delta\mathbf p_{\rm prog}
&:= a_\parallel \operatorname{sinc}\!\left(\frac{a_\psi}{2}\right)
\mathbf u_\parallel(\bar{\psi}),\\
\Delta\mathbf p_{\rm swing}
&:= 2r\sin\!\left(\frac{a_\psi}{2}\right)
\mathbf u_\perp(\bar{\psi}).
\end{aligned}
\label{eq:prelim_nonholonomic_update}
\end{equation}
Here, $\Delta\mathbf p_{\rm prog}$ is the forward progress of the
no-slip reference point, and $\Delta\mathbf p_{\rm swing}$ is the
signed box-center swing caused by rotating the offset $r$. Since
$\Delta\mathbf p_{\rm prog}$ is parallel to
$\mathbf u_\parallel(\bar{\psi})$, projecting Eq.~\eqref{eq:prelim_nonholonomic_update}
onto $\mathbf u_\perp(\bar{\psi})$ leaves only the swing term:
\begin{equation}
\Delta\mathbf p_{\rm nh}^{\top}\mathbf u_\perp(\bar{\psi})
=
2r\sin\!\left(\frac{a_\psi}{2}\right).
\label{eq:prelim_nonholonomic_lateral}
\end{equation}

\subsection{Closed-Loop Covariate Shift}
\label{subsec:cl_covariate_shift}
Behavior cloning (BC) on a logged dataset $\mathcal D=\{(\mathcal H_0,\tau)\}$ minimizes the negative log-likelihood for model parameters~$\theta$:
\begin{equation}
\mathcal{L}_{\mathrm{BC}}(\theta)
=
-\,\mathbb{E}_{\tau \sim \mathcal{D}}
\sum_{t=0}^{T-1}
\log p_\theta\!\left(\mathbf s_{t+1}\,\middle|\,\mathcal H_t\right).
\label{eq:bc_loss}
\end{equation}
This is teacher forcing~\cite{williams1989learning}: the conditioning context $\mathcal H_t$ is drawn from logged data. At deployment, however, the simulator conditions on its own previous outputs,
\begin{equation}
\hat{\mathbf s}_{t+1}
\sim
p_\theta\!\left(\,\cdot\,\middle|\,\hat{\mathcal H}_t\right),
\qquad t=0,\ldots,T-1,
\label{eq:cl_rollout}
\end{equation}
where $\hat{\mathcal H}_t$ is built from the initial history and generated states. The induced closed-loop rollout distribution $p_\theta^{\mathrm{CL}}(\tau\mid\mathcal H_0)$ can differ from the data distribution $p_{\mathrm{data}}(\tau\mid\mathcal H_0)$ because the model sees off-data contexts created by its own predictions, causing compounding error and closed-loop covariate shift~\cite{ross2011reduction}.
One remedy is to align closed-loop rollouts with the data distribution,
    $p_\theta^{\mathrm{CL}}(\tau\mid\mathcal H_0)\approx p_{\mathrm{data}}(\tau\mid\mathcal H_0)$, for example by minimizing a reverse-KL objective, $D_{\mathrm{KL}}(p_\theta^{\mathrm{CL}}\|p_{\mathrm{data}})$~\cite{ke2020imitation} that penalizes generated rollouts unlikely under the data distribution. However, reverse-KL is mode-seeking~\cite{ke2020imitation} and can concentrate mass on a few high-density modes. 

\subsection{Training Flow-based Model}
\label{subsec:prelim_fm}
Flow Matching~\cite{lipman2022flow} defines a probability path $(p_\lambda)_{0\le\lambda\le1}$ from a Gaussian source $p_0=\mathcal N(\mathbf 0,\mathbf I)$ to a data target $p_1=p_{\mathrm{data}}$ over $\mathbf x\in\mathbb R^d$. It learns a velocity field $\mathbf v:[0,1]\times\mathbb R^d\rightarrow\mathbb R^d$ whose flow map $\varphi:[0,1]\times\mathbb R^d\rightarrow\mathbb R^d$ satisfies
\begin{equation}
\frac{d}{d\lambda}\varphi_\lambda(\mathbf{x})
=
\mathbf v\!\left(\lambda,\varphi_\lambda(\mathbf{x})\right),
\qquad
\varphi_0(\mathbf{x})=\mathbf{x}.
\label{eq:fm_flow_map}
\end{equation}
Given a clean data sample $\mathbf x_1$ and noise $\mathbf x_0\sim p_0$, we use the affine optimal-transport path
$\mathbf{x}_\lambda
=
(1-\lambda) \mathbf{x}_0+\lambda\mathbf{x}_1.
$

Corresponding velocity target
$\mathbf v^\star
=
\mathbf{x}_1-\mathbf{x}_0,$
and the encoded context $\mathbf e$, derives our flow-matching loss:
\begin{equation}
\resizebox{0.98\columnwidth}{!}{$\displaystyle
\mathcal{L}_{\mathrm{FM}}(\theta)
=
\mathbb{E}_{(\mathbf{x}_1,\mathbf e)\sim\mathcal{D},\,\mathbf{x}_0\sim p_0,\,\lambda\sim \mathcal{U}[0,1]}
\left[
\left\|
\mathbf v_\theta(\mathbf{x}_\lambda,\lambda,\mathbf e)
-
\mathbf v^\star
\right\|_2^2
\right],
$}
\label{eq:fm_loss}
\end{equation}
where $\mathbf{v}_\theta$ is a parameterized neural velocity field. After training, samples are generated by solving the learned ODE:
\begin{equation}
\frac{d\mathbf{x}_\lambda}{d\lambda}
=
\mathbf v_\theta(\mathbf{x}_\lambda,\lambda,\mathbf e),
\qquad
\mathbf{x}_0\sim p_0.
\label{eq:fm_ode}
\end{equation}

\section{Method}
\label{sec:method}

\begin{figure*}[t]
  \centering
  \includegraphics[width=0.94\linewidth]{\FinalRoot 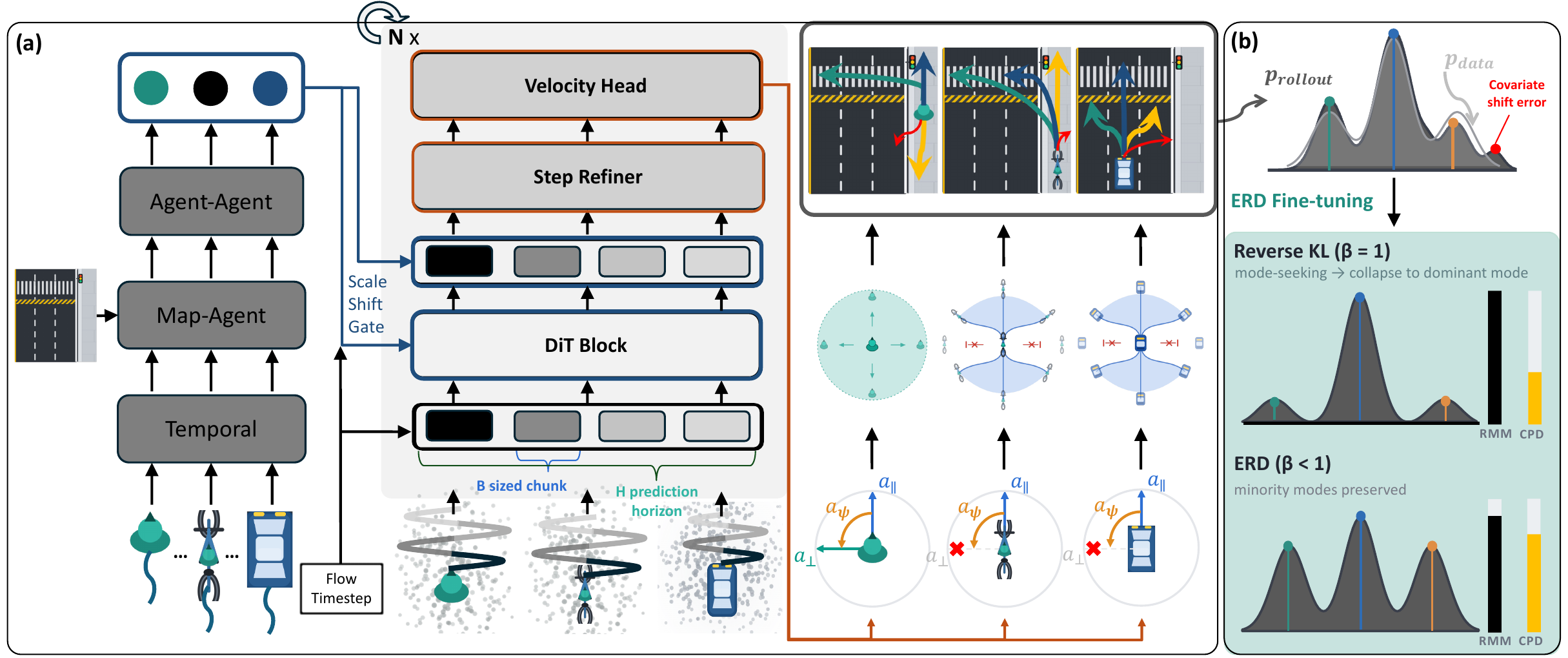}
  \caption{Overview of Flow-ERD. (a) The Agent-Type Aware Flow-Matching (AFM) backbone generates a shared continuous action representation, executed through type-specific kinematics (non-holonomic for vehicles/cyclists, holonomic for pedestrians). 
  (b) Entropy-Regularized Distillation (ERD) then fine-tunes the closed-loop distribution: the vanilla reverse-KL objective ($\beta = 1$) is mode-seeking, easier to collapse onto the dominant (straight) mode, whereas ERD ($\beta < 1$) targets an entropy-tempered distribution that preserves minority modes while still mitigating closed-loop covariate shift.}
  \label{fig:method}
  \vspace{-15pt}
\end{figure*}

In this section, we introduce our two components: the proposed backbone, Agent-Type Aware Flow Matching (AFM), and its diversity-preserving fine-tuning method, Entropy-Regularized Distillation (ERD), illustrated in \cref{fig:method}. AFM captures multimodal diversity while keeping it realistic through agent-type aware modeling in a continuous action space. ERD then fine-tunes this with an entropy-regularized objective that retains minority modes, avoiding the mode collapse of standard reverse KL objectives.

\subsection{Flow-Matching Backbone}
\label{sec:afm}
\subsubsection{Motivation and Design Rationale}
Token-based simulators execute discrete motion primitives, but finite vocabularies limit continuous variation. Continuous generators remove this codebook bottleneck, but without type-aware execution, they may assign probability to motions invalid for a given agent type; in closed loop, such invalid states become future context. AFM separates generation from execution: the flow model samples multimodal continuous kinematic actions, and the simulator feeds back only poses obtained by executing those actions through type-specific transitions.

\subsubsection{Kinematic Action-Space Flow Modeling}
For agent $i$ at time $t$, we use the metric kinematic action
\begin{equation}
\mathbf a_t^i=(a_{\parallel,t}^i,a_{\perp,t}^i,a_{\psi,t}^i)\in\mathcal A_{c_i},
\label{eq:afm_action}
\end{equation}
where the entries are local longitudinal, lateral, and heading increments. Given the history context $\mathcal H_t$, AFM models the full $H$-step action sequence
$\mathbf a_{t:t+H-1}$. In flow-matching pretraining, the clean endpoint $\mathbf{x}_1$ becomes the target action sequence $\mathbf a_{t:t+H-1}^\star$; we define $\mathbf a_{t:t+H-1}^\star$ below after specifying the transition that executes actions into poses.

\subsubsection{Type-Specific State Transition}
To map an action to the next pose, AFM applies the agent-type-specific transition $\mathcal F_{c_i}$ to the planar state $\mathbf s_t^i=(\mathbf p_t^i,\psi_t^i)$:
\begin{equation}
\begin{aligned}
\mathbf s_{t+1}^i &= \mathcal F_{c_i}(\mathbf s_t^i,\mathbf a_t^i),\\
\text{with}\quad
\psi_{t+1}^i &= \operatorname{wrap}(\psi_t^i+a_{\psi,t}^i),\\
\mathbf p_{t+1}^i &= \mathbf p_t^i+\Delta\mathbf p_t^i,
\end{aligned}
\label{eq:afm_forward}
\end{equation}
where $\operatorname{wrap}$ maps angles to $[-\pi,\pi]$. Thus, $\mathcal F_{c_i}$ is fully specified by the heading update and the type-dependent displacement below. The map-frame displacement uses the motion models in \cref{subsec:transition}:
\begin{equation}
\Delta\mathbf p_t^i=
\begin{cases}
\Delta\mathbf p_{\mathrm{hol}}^i, & c_i=\textsc{ped},\\
\Delta\mathbf p_{\mathrm{nh}}^i, & c_i\in\{\textsc{veh},\textsc{cyc}\}.
\end{cases}
\label{eq:afm_transition}
\end{equation}
For vehicles and cyclists, the non-holonomic branch uses forward displacement, heading change, and the no-slip offset $r_i$; the lateral channel $a_{\perp,t}^i$ remains in the shared action vector but is not used in their pose update. This preserves a shared action space while enforcing type-compatible execution.

\subsubsection{Training: Transition-Consistent Action Targets}
\label{subsubsec:afm_targets}
To supervise AFM in action space, logged pose pairs must be
converted into actions. For vehicles and cyclists, this requires
inverting Eq.~(2), which depends on the no-slip offset $r$.
The dataset logs label only box-center poses, not the no-slip
point, so this offset must be estimated from the logged
trajectories. We set $r_i=\rho_{c_i}\ell_i$, where $c_i$ is agent $i$'s
type and $\ell_i$ its box length, \textcolor{blue}{which gives \cref{eq:prelim_nonholonomic_lateral} into:}
\begin{equation}
\hat{\rho}_{i}
=
\frac{
\Delta\mathbf p_i^\top \mathbf u_\perp(\bar{\psi}_i)
}{
2\ell_i\sin(a_{\psi,i}/2)
}.
\label{eq:afm_rho}
\end{equation}

\textcolor{blue}{
We retain only intervals with $|2\ell_i\sin(a_{\psi,i}/2)|\ge\delta$, where $ \delta = 0.10\mathrm{m}$, excluding
ill-conditioned cases, and aggregate the retained estimates by type with a
robust weighted median, yielding one $\rho_c$ per non-holonomic type.}

With the no-slip ratio fixed, we convert logged poses into an
executable action sequence. During inference, actions are executed
sequentially, and each executed pose becomes the state for the
next action. To reduce the train-inference gap, we construct the targets in the same way: rather than starting from ground truth position $\mathbf{s}_{t}^{i,\star}$, we initialize previously executed state $\tilde{\mathbf s}_{t}^{i}$ for $k=0,\ldots, H-1$, then recover the action toward the next logged pose:
\begin{equation}
\mathbf a_{t+k}^{i,\star}
=
\mathcal F_{c_i}^{-1}
(\tilde{\mathbf s}_{t+k}^{i},\mathbf s_{t+k+1}^{i,\star}),
\label{eq:afm_inverse}
\end{equation}
where $\mathcal F_{c_i}^{-1}$ inverts $\mathcal F_{c_i}$.
We then execute this action and use the resulting pose as the
state for the next step sequentially:
\begin{equation}
\tilde{\mathbf s}_{t+k+1}^{i}
=
\mathcal F_{c_i}(\tilde{\mathbf s}_{t+k}^{i},\mathbf a_{t+k}^{i,\star}).
\label{eq:afm_rolling}
\end{equation}
This makes $\mathbf a_{t,0:H-1}^{\star}$ an executable
flow-matching target whose state sequence follows the same
transition used at inference. We filter out the targets with large re-execution error $\left|\tilde{\mathbf{s}}^i_{t+k+1}-\mathbf{s}_{t+k+1}^{i,\star}\right|$ to improve training stability.

\subsubsection{Closed-Loop Inference}
At inference, AFM generates a full $H$-step action sequence. The simulator executes it with $\mathcal F_{c_i}$, commits only the first $B<H$ steps in a receding horizon manner, and appends their poses to the next context as in \cref{eq:cl_rollout}. \textcolor{blue}{In all experiments, $H{=}20$ ($2$\,s at $10$\,Hz) and $B{=}5$ ($0.5$\,s), and \cref{eq:fm_ode} is integrated with $16$ Euler steps, so an $8$\,s rollout takes $16$ replanning steps.} We denote this receding-horizon closed-loop rollout distribution as $p_\theta^\text{CL}(\tau|\mathcal{H}_0)$ and the $H$-step direct prediction distribution as $p_\theta^\text{OL}(\tau|\mathcal{H}_0)$. Under $p_\theta^\text{CL}(\tau|\mathcal{H}_0)$, since feedback uses type-specific executed poses rather than unconstrained pose predictions, AFM preserves continuous multimodality while reducing type-incompatible closed-loop states.

\subsubsection{Architecture}
Fig.~\ref{fig:method} summarizes the AFM architecture. Given $\mathcal H_t$, a SMART-style scene encoder builds temporal, map--agent, and agent--agent context features, summarized as $\mathbf e_t=\mathrm{Enc}(\mathcal H_t)$. The temporal history is encoded with continuously executed motion features rather than discrete motion tokens.

The flow decoder takes $\mathbf{e_t}$, a noisy $H$-step action sequence $\mathbf{x_\lambda}$, with flow noise timestep $\lambda$. It embeds the sequence into $B$-step chunks and applies chunk-level self-attention with DiT-style scale--shift--gate conditioning~\cite{peebles2023scalable}. A step refiner applies within-chunk self-attention, and an MLP velocity head outputs the action-space velocity field $\mathbf {v} _\theta(\mathbf{x}_\lambda,\lambda,\mathbf{e}_t)$ used in Eq.~\eqref{eq:fm_ode}. The resulting action sequence is executed by the type-specific transition in Eq.~\eqref{eq:afm_transition}. \textcolor{blue}{The generator has $7.12$M parameters.}

\begin{algorithm}[t]
\caption{Entropy-Regularized Distillation (ERD)}
\label{alg:erd}
\begin{algorithmic}[1]
\REQUIRE pretrained $\theta_0$; dataset $\mathcal{D}$; temperature $\beta$; critic steps $n_{\mathrm{critic}}$; phase length $L_{\mathrm{phase}}$; fake-score step size $\eta$; stop-gradient operator $\mathrm{sg}$.
\STATE initialize generator $\theta$ and fake-score parameters $\phi$ from $\theta_0$; freeze real score $\mathbf g_{\theta_0}^{\mathrm{OL}}$
\REPEAT
  \FOR{$q = 1$ \textbf{to} $L_{\mathrm{phase}}$ \hfill $\triangleright$ \textsc{Fake only $\phi$}}
    \STATE \finalchange{Sample $\mathcal H_0$; generate the initial closed-loop segment $\hat{\mathbf a}_{0:H-1}$.}
    \STATE $\phi \leftarrow \phi-\eta\nabla_\phi \mathcal L_{\mathrm{FM}}(\phi;\mathbf{x}_1=\mathrm{sg}[\hat{\mathbf a}_{\finalchange{0:H-1}}])$
  \ENDFOR
  \FOR{$q = 1$ \textbf{to} $L_{\mathrm{phase}}$ \hfill $\triangleright$ \textsc{Both $\theta$, $\phi$}}
    \STATE \finalchange{Sample $\mathcal H_0$; generate the initial closed-loop segment $\hat{\mathbf a}_{0:H-1}$.}
    \STATE $\phi \leftarrow \phi-\eta\nabla_\phi \mathcal L_{\mathrm{FM}}(\phi;\mathbf{x}_1=\mathrm{sg}[\hat{\mathbf a}_{\finalchange{0:H-1}}])$
    \STATE \textbf{if} $q \bmod n_{\mathrm{critic}} = 0$\,\textbf{:} \; \finalchange{update $\theta$ via \cref{eq:erd_grad_final} on this initial $H$-step segment.}
  \ENDFOR
\UNTIL{converged}
\RETURN $\theta$
\end{algorithmic}
\end{algorithm}

\subsection{Entropy-Regularized Distillation}
\label{subsec:erd}
We fine-tune the AFM backbone under closed-loop rollouts to mitigate
covariate shift, while explicitly preserving the diversity it already
represents. \textcolor{blue}{
Motivated by the distribution-matching goal $p_\theta^{\mathrm{CL}}(\tau\mid \mathcal H_0)
\approx p_{\mathrm{data}}(\tau\mid \mathcal H_0)$ in \cref{subsec:cl_covariate_shift}, we define ERD directly over the executable AFM action rollouts, which deterministically induce the closed-loop state evolution.
Let $\mathbf A=(\mathbf a_0,\ldots,\mathbf a_{T-1})$ denote the action rollout over the simulation horizon. We define the reverse-KL objective as:}
\begin{equation}
J(\theta)
=
\mathbb{E}_{\mathcal H_0\sim\mathcal D}
\left[
D_{\mathrm{KL}}
\left(
\textcolor{blue}{p_\theta^{\mathrm{CL}}(\mathbf A\mid\mathcal H_0)}
\,\middle\|\,
\textcolor{blue}{p_{\mathrm{data}}(\mathbf A\mid\mathcal H_0)}
\right)
\right].
\label{eq:rkl}
\end{equation}
Reverse-KL is mode-seeking in practice: it penalizes model rollouts in low-density regions of \(p_{\mathrm{data}}\), but because the expectation is over \(p_\theta^{\mathrm{CL}}\), valid data modes rarely sampled by the model contribute little to the loss. 
Under imperfect capacity or optimization, probability can therefore
concentrate on dominant modes and underrepresent minority ones.
To counteract this, we add an entropy regularizer:
\begin{equation}
\resizebox{0.98\columnwidth}{!}{$\displaystyle
J_{\mathrm{ERD}}(\theta)
=
\mathbb{E}_{\mathcal H_0}\!\Bigl[
D_{\mathrm{KL}}\!\bigl(p_\theta^{\mathrm{CL}}\,\big\|\,p_{\mathrm{data}}\bigr)
-\gamma\,\operatorname{Ent}\!\bigl(
\textcolor{blue}{p_\theta^{\mathrm{CL}}(\mathbf A\mid\mathcal H_0)}
\bigr)
\Bigr],
$}
\label{eq:erd_obj}
\end{equation}
where $\operatorname{Ent}(\cdot)$ denotes Shannon entropy and $\gamma\ge 0$
is its regularization weight. For brevity, we hereafter use the temperature
$\beta := 1/(1+\gamma)\in(0,1]$.

Expanding the entropy and dividing by $1+\gamma$ shows that, up to a positive scale and
an additive constant, \cref{eq:erd_obj} is an ordinary reverse-KL divergence against a
tempered target,
\begin{equation}
J_{\mathrm{ERD}}(\theta) \;=\; (1+\gamma)\,\mathbb{E}_{\mathcal H_0}\!\left[\,D_{\mathrm{KL}}\!\bigl(p_\theta^{\mathrm{CL}}\,\big\|\,p_{\mathrm{data}}^{\beta}\bigr)\right] + \mathrm{const},
\label{eq:erd_equiv}
\end{equation}
\begin{equation}
    p_{\mathrm{data}}^{\beta} \propto
    \textcolor{blue}{p_{\mathrm{data}}(\mathbf A\mid \mathcal H_0)^{\beta}}.
\end{equation}
\finalchange{The full-rollout objective} is therefore plain distribution matching toward $p_{\mathrm{data}}^{\beta}$, whose
\finalchange{distributional} minimizer is the tempered fixed point $p_\theta^{\mathrm{CL}\star}\!\propto\!p_{\mathrm{data}}^{\beta}$.
Tempering reduces the density contrast between modes and relaxes the mode-dropping bias of
reverse-KL: smaller $\beta$ (larger $\gamma$) flattens $p_{\mathrm{data}}^{\beta}$ toward a
uniform distribution over the data support and up-weights minority modes, while
$\beta\!=\!1$ ($\gamma\!=\!0$) recovers plain distribution matching toward $p_{\mathrm{data}}$.

\begin{table*}[t]
\centering
\caption{WOSAC 2025 test split. Best in \textbf{bold}, second best \underline{underlined}, $\dagger$: fine-tuned from SMART.}
\label{tab:main_wosac}
\setlength{\tabcolsep}{4pt}
\begin{tabular}{c l c c c c c}
\toprule
\textbf{Type} & \textbf{Method} & \textbf{RMM}~$\uparrow$ & \textbf{Kin.}~$\uparrow$ & \textbf{Int.}~$\uparrow$ & \textbf{Map}~$\uparrow$ & \textbf{minADE}~$\downarrow$ \\
\midrule

\multirow{5}{*}{\rotatebox[origin=c]{90}{\textbf{Backbone}}}
  & SMART~\cite{wu2024smart}                                  & 0.7814 & 0.4854 & 0.8089 & 0.9153 & 1.3931 \\
  & TrajTok~\cite{zhang2026trajtok}                               & \textbf{0.7852} & 0.4887 & \textbf{0.8116} & \textbf{0.9207} & 1.3179 \\
  & UniMM~\cite{lin2025revisitmixturemodelsmultiagent}           & 0.7829 & 0.4914 & 0.8089 & 0.9161 & \textbf{1.2949} \\
  & MDG~\cite{huang2025mdg}                                       & 0.7844 & \underline{0.4928} & \underline{0.8099} & \underline{0.9183} & \underline{1.3123} \\
  & \textbf{AFM backbone }(ours)                          & \underline{0.7845}    & \textbf{0.4955}    & 0.8094    & 0.9178    & 1.3358 \\
\midrule
\multirow{6}{*}{\rotatebox[origin=c]{90}{\textbf{Fine-tuned}}}
  & CAT-K\textsuperscript{$\dagger$}~\cite{zhang2025closed}                      & 0.7846 & 0.4931 & 0.8106 & 0.9177 & 1.3065 \\
  & RoaD\textsuperscript{$\dagger$}~\cite{garcia2025road}                                    & 0.7847 & 0.4932 & 0.8106 & 0.9178 & 1.3042 \\
  & RLFTSim\textsuperscript{$\dagger$}~\cite{ahmadi2026rlftsim}                              & 0.7857 & 0.4927 & \underline{0.8129} & 0.9183 & 1.3252 \\
  & SMART-R1\textsuperscript{$\dagger$}~\cite{pei2026advancing}                             & 0.7858 & \underline{0.4944} & 0.8110 & \textbf{0.9201} & \underline{1.2885} \\
  & DecompGAIL\textsuperscript{$\dagger$}~\cite{guo2026decompgail}                          & \underline{0.7864} & 0.4919 & \textbf{0.8152} & 0.9176 & 1.4209 \\
  & \textbf{Flow-ERD} (AFM + ERD)                                & \textbf{0.7878}    & \textbf{0.5062}    & 0.8110    & \underline{0.9190}    & \textbf{1.2721} \\
\bottomrule
\end{tabular}
\end{table*}
\begin{table*}[t]
\centering
\caption{WOSAC 2025 4\% validation split. Best in \textbf{bold}, second best \underline{underlined}, $\dagger$: fine-tuned from SMART.
}
\label{tab:val_cpd}
\setlength{\tabcolsep}{4pt}
\begin{tabular}{c l c c c c c c}
\toprule
\textbf{Type} & \textbf{Method} & \textbf{RMM}~$\uparrow$ & \textbf{Kin.}~$\uparrow$ & \textbf{Int.}~$\uparrow$ & \textbf{Map}~$\uparrow$ & \textbf{minADE}~$\downarrow$ & \textbf{CPD}~$\uparrow$ \\
\midrule
\multirow{6}{*}{\rotatebox[origin=c]{90}{\textbf{Backbone}}}
  & SMART~\cite{wu2024smart}                           & 0.7807 & \underline{0.4873} & 0.8071 & 0.9144 & 1.323 & \underline{0.1655} \\
  & TrajTok~\cite{zhang2026trajtok}                    & \textbf{0.7837} & 0.4851 & \textbf{0.8110} & \textbf{0.9193} & 1.3258 & 0.1587 \\
  & UniMM~\cite{lin2025revisitmixturemodelsmultiagent} & 0.7812 & 0.4863 & 0.8070 & 0.9165 & \textbf{1.2725} & 0.1514 \\
& \textbf{AFM backbone (ours)}                & \underline{0.7836} & \textbf{0.4914} & \underline{0.8096} & \underline{0.9172} & \underline{1.3024} & \textbf{0.1858} \\ 
  \cmidrule(l){2-8}
  & All holonomic FM ablation                          & 0.7823 & 0.4905 & 0.8063 & 0.9182 & 1.2993 & 0.2046 \\
  & All non-holonomic FM ablation                      & 0.7824 & 0.4910 & 0.8091 & 0.9144 & 1.2960 & 0.1843 \\

\midrule
\multirow{7}{*}{\rotatebox[origin=c]{90}{\textbf{Fine-tuned}}}
  & CAT-K~\cite{zhang2025closed}\textsuperscript{$\dagger$}       & 0.7842 & 0.4904 & 0.8110 & 0.9175 & 1.3001 & 0.1491 \\
  & RoaD~\cite{garcia2025road}\textsuperscript{$\dagger$}         & 0.7847 & 0.4904 & \underline{0.8116} & 0.9182 & \textbf{1.2603} & 0.1585 \\
  & RLFTSim~\cite{ahmadi2026rlftsim}\textsuperscript{$\dagger$}   & 0.7848 & 0.4895 & \underline{0.8116} & \textbf{0.9190} & 1.2988 & 0.1510 \\
  & DecompGAIL~\cite{guo2026decompgail}\textsuperscript{$\dagger$} & 0.7854 & 0.4912 & \textbf{0.8123} & \textbf{0.9190} & 1.2868 & 0.1420 \\
  & RoaD (AFM Backbone)
  & 0.7856
  & 0.5001
  & 0.8102
  & 0.9172
  & 1.2926
  & 0.1639 \\
  & \textbf{Flow-ERD} ($\beta=1.0$)         & \textbf{0.7876} & \textbf{0.5063} & 0.8108 & 0.9184 & \underline{1.2620} & \underline{0.1684} \\
  & \textbf{Flow-ERD} ($\beta=0.99$)        & \underline{0.7869} & \underline{0.5053} & 0.8099 & 0.9182 & 1.2723 & \textbf{0.1828} \\

\bottomrule
\vspace{-20pt}
\end{tabular}
\end{table*}
The likelihood $\log p_\theta^{\mathrm{CL}}$ is unavailable in closed form for autoregressive
flow rollouts, so we cannot optimize \cref{eq:erd_equiv} directly. \finalchange{In practice, we use}
Distribution-Matching Distillation (DMD)~\cite{yin2024one} on the model's generated action
sequences, following Self-Forcing~\cite{huang2025selfforcing}.
\finalchange{This is a local surrogate of \cref{eq:erd_equiv} on the initial $H$-step closed-loop segment.}

Let \(\hat{\mathbf x}_1 \equiv \hat{\mathbf a}_{\finalchange{0:H-1}}\) denote the generated clean \(H\)-step kinematic-action sequence, and \(\hat{\tau}\) be the state rollout induced by executing \(\hat{\mathbf a}_{\finalchange{0:H-1}}\) through the type-specific transitions \(\mathcal F=\{\mathcal F_c\}_{c\in\mathcal C}\).
DMD sidesteps the intractable likelihood by expressing the KL gradient through scores $\mathbf g_p(\mathbf{x})=\nabla_{\mathbf{x}}\log p(\mathbf{x})$ of the
two action-sequence distributions, evaluated at each flow noise timestep $\lambda$.
\finalchange{Here, $p_{\mathrm{data}}$ denotes the $H$-step data distribution given $\mathcal H_0$, and the standard DMD real score is replaced by that of its tempered target,}
$\mathbf g_{p_{\mathrm{data}}^{\beta}}=\beta\mathbf g_{\mathrm{p_\mathrm{data}}}$, where $\mathbf g_{p_\mathrm{data}}$ is the data score.
We make this practical with two substitutions: we use the frozen pretrained backbone
$\mathbf g_{\theta_0}^{\mathrm{OL}}$ as the data score, where OL denotes the $H$-step action distribution learned under logged-state conditioning during open-loop pretraining. Since $p_{\theta_0}^{\mathrm{OL}}\!\approx\!p_{\mathrm{data}}$
on the data support~\cite{huang2025selfforcing}, the real score is evaluated from this frozen OL model, while the CL fake score is trained on action sequences generated by rolling the current policy forward with $B$-step commitment. \finalchange{Both scores are conditioned on the same initial context $\mathcal H_0$, omitted below; intermediate closed-loop contexts are used only by the generator.} We realize the tempered score by scaling
it directly, $\mathbf g_{p_{\mathrm{data}}^{\beta}}\!\approx\!\beta\,\mathbf g_{\theta_0}^{\mathrm{OL}}$; \finalchange{the underlying identity is exact for true scores}
at the clean-data end $\lambda\!\to\!1$, as in classifier-free
guidance~\cite{ho2022classifierfreediffusionguidance}.
\finalchange{This yields our surrogate for $\nabla_\theta J_{\mathrm{ERD}}$,}
\begin{equation}
\finalchange{\mathbf G_{\mathrm{ERD}}^{(0:H)}(\theta)}
\propto
\mathbb{E}_{\hat{\mathbf x}_\lambda,\,\lambda}
\!\left[
\left(
\mathbf g_\phi^{\mathrm{CL}}(\hat{\mathbf x}_\lambda,\lambda)
-\beta\,\mathbf g_{\theta_0}^{\mathrm{OL}}(\hat{\mathbf x}_\lambda,\lambda)
\right)
\partial_\theta\hat{\mathbf x}_1
\right],
\label{eq:erd_grad_final}
\end{equation}
where $\hat{\mathbf x}_\lambda$ is $\hat{\mathbf x}_1$ noised to flow time $\lambda$ as in the affine path above
and the fake score $\mathbf g_\phi^{\mathrm{CL}}$ is kept on-policy. Detailed algorithm can be found in~\cref{alg:erd}.  \textcolor{blue}{In practice, for vehicles and cyclists, the non-executable $a_\perp$ channel is masked from the DMD update and surrogate loss, while pedestrians use all three action dimensions.}

\section{Experiments}
\label{sec:exp}
In this section, we evaluate our framework, Flow-ERD, around three questions. (1) Does Agent-Type Aware Flow Matching (AFM) achieve higher realism while maintaining high diversity? (2) Does Entropy-Regularized Distillation (ERD) preserve this diversity while improving realism? (3) Does the preserved diversity translate into semantic multimodality with minority modes retained? To address these questions, we first introduce our evaluation metrics and then present our analysis.

\begin{figure}[t]
  \centering
  \includegraphics[width=\linewidth,trim=0pt 5pt 0pt 5pt,clip]{\FinalRoot 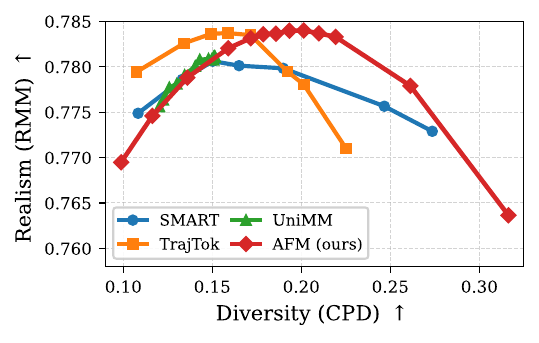}
  \caption{Realism--diversity trade-off on the validation split. UniMM, SMART, and TrajTok sweep $k$ of top-k decoding during validation rollouts, whereas AFM (ours) sweeps the Gaussian noise scale. AFM traces the upper-right Pareto frontier, reaching an RMM of 0.7840 at noise scale 1.05.
  }
  \label{fig:pareto}
  \vspace{-20pt}
  
\end{figure}

\begin{figure}[t]
  \centering
  \includegraphics[width=\linewidth,trim=0pt 5pt 0pt 5pt,clip]{\FinalRoot 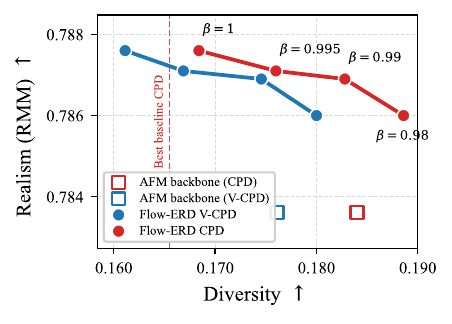}
  \caption{\textcolor{blue}{ERD entropy temperature $\beta$ sweep on the validation split: realism (RMM) versus total diversity (CPD, \cref{eq:cpd}) and \finalchange{vehicle-}validity-gated diversity (V-CPD, \cref{subsec:main_results}); open squares mark the pretrained backbone, and the dashed line the best baseline CPD (SMART, \cref{tab:val_cpd}).
  }
  }
  \label{fig:beta_sweep}
  \vspace{-10pt}

\end{figure}

\begin{figure}[t]
  \centering
  \includegraphics[width=\columnwidth,trim=0pt 7pt 0pt 4pt,clip]{\FinalRoot 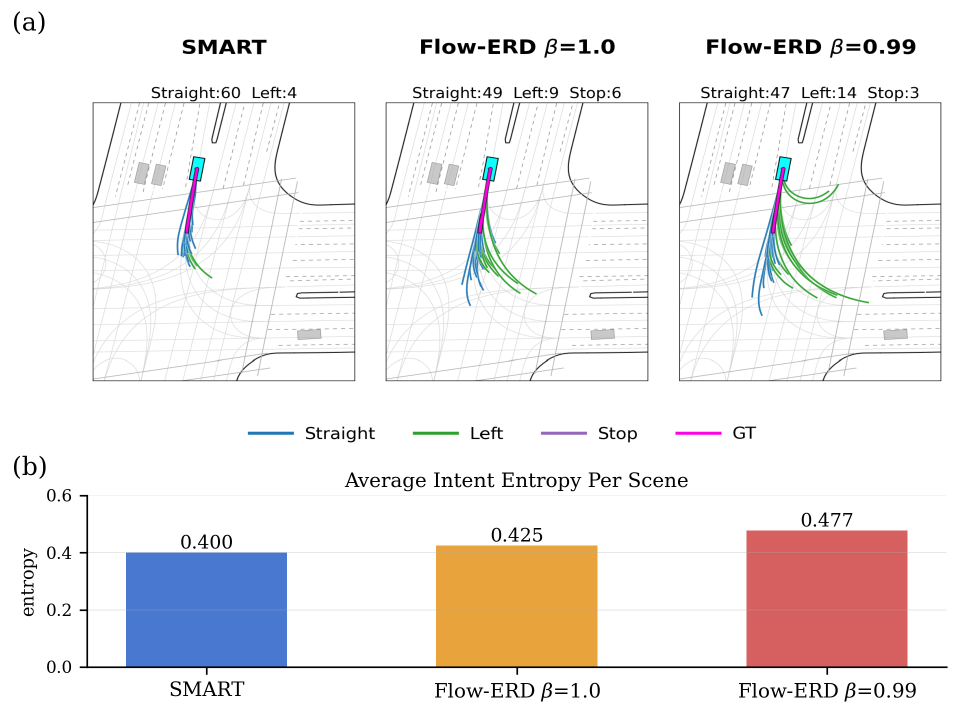}
  \caption{We run multi-agent closed-loop rollouts over 1,048 validation scenes and label ego-maneuver intents following WOMD~\cite{Ettinger_2021_ICCV}. \textbf{(a)} Ego-trajectory diversity on a WOSAC scene, shown by overlaying closed-loop rollouts.  \textbf{(b)} Average per-scene intent entropy of the ego rollouts.}
  \vspace{-20pt}
  \label{fig:qualitative}
\end{figure}

\subsection{Evaluation Metrics} 
\label{subsec:metrics}
\paragraph{Realism}
We follow the WOSAC 2025 evaluation~\cite{montali2023waymo}, whose realism meta-metric (RMM) aggregates distributional likelihoods over kinematic, interactive, and map-based statistics against the single logged future. We report RMM and its three components, and refer to~\cite{montali2023waymo} for exact definitions. We additionally report minADE (per-object minimum ADE over rollouts), which does not enter RMM but is standard for behavior-prediction comparison.
\paragraph{Diversity}
RMM is a likelihood-based score against a single logged future, so it credits spread around the log but misses plausible modes the log omits. This cannot distinguish dominant-mode tracking from realistic yet diverse rollouts. Therefore, we define the log-independent diversity metric.

Concretely, given $K$ closed-loop rollouts of the same scenario, we define the type-normalized pairwise distance:
\begin{equation}
d(\tau, \tau') \;=\; \sqrt{\,\sum_{c \in \mathcal{C}} \frac{1}{\sigma_c^{2}} \cdot \frac{1}{N_cT}\!\!\sum_{\substack{i:\,c_i=c\\1\le t \le T}}\!\!\bigl\|\mathbf p^i_t - \mathbf p'^{\,i}_t\bigr\|^{2}\,},
\label{eq:cpd_pair}
\end{equation}
where $\mathbf p^i_t, \mathbf p'^{\,i}_t \in \mathbb{R}^2$ are agent $i$'s positions at step $t$ in the two rollouts, $\sigma_c$ is a per-type scale fixed on the training set, and $N_c$ is the number of agents of type $c$; types with $N_c=0$ are omitted. The Cross-Pair Diversity (CPD) averages \cref{eq:cpd_pair} over all unordered rollout pairs and scenarios $\omega \in W$:
\begin{equation}
\begin{aligned}
\mathrm{CPD}
&=
\frac{1}{|\mathcal{W}|}\sum_{\omega\in \mathcal{W}}
\frac{2}{K(K-1)}
\sum_{1 \le k < k' \le K}
d\bigl(\tau_\omega^{(k)}, \tau_\omega^{(k')}\bigr).
\end{aligned}
\label{eq:cpd}
\end{equation}
\textcolor{blue}{Higher CPD indicates broader rollout spread but cannot distinguish genuine multimodality from prediction error or closed-loop drift~\cite{suo2021trafficsim}. We compare CPD at matched RMM and examine its validity and intent-level structure in~\cref{subsec:main_results}.}

\subsection{Experimental Results}
\label{subsec:main_results}

\subsubsection{AFM breaks the backbone realism–diversity trade-off}
On the WOSAC test split (\cref{tab:main_wosac}), the AFM backbone attains the highest overall RMM among continuous backbones and is competitive with the strongest tokenized backbones such as TrajTok. Without any fine-tuning, it achieves the best kinematic component among all baselines, including fine-tuned, confirming that agent-type aware kinematic modeling most faithfully reproduces per-type kinematics.

Its advantage is clearest on diversity (\cref{tab:val_cpd}): across all other pretrained baselines, AFM attains the highest CPD. To rule out that this merely reflects suboptimal sampling settings for the baselines, we sweep each model's controllable knob in \cref{fig:pareto}: top-$k$ token/anchor selection for token-based models and UniMM, and the Gaussian noise scale applied to $x_0\sim p_0$ in \textcolor{blue}{\cref{eq:fm_ode}} for ours; the baselines cannot reach our frontier under any setting. In most cases, AFM is higher in both RMM and CPD, and against TrajTok, whose RMM matches ours, AFM still attains substantially higher CPD. At matched realism, this gap cannot \textcolor{blue}{solely} stem from drift-induced variance, and instead shows AFM placing mass on genuinely more diverse futures, RMM alone would not reveal. The gap to UniMM indicates this diversity stems not only from a continuous representation or higher resolution, but also from the expressiveness of our flow-based modeling.

\Cref{tab:val_cpd} ablates whether continuous flow matching alone suffices, or whether the transition must be agent-type aware. The all-holonomic backbone yields the largest raw CPD but the lowest kinematic score, indicating that much of its spread comes from type-incompatible motion freedom, lateral slip for vehicles and cyclists. Enforcing a non-holonomic transition for all agents recovers kinematic realism and trims this unrealistic spread, but it over-constrains pedestrians, yielding lower RMM and CPD than AFM. AFM resolves both failure modes by assigning holonomic transitions to pedestrians and non-holonomic ones to vehicles and cyclists, achieving the best RMM and kinematic score among the other variants.

\textit{2) ERD improves closed-loop realism without collapsing diversity:}
\Cref{tab:main_wosac} shows that Flow-ERD achieves the best RMM among all baselines. We report the official submission result with the vanilla reverse-KL objective ($\beta=1.0$), since the leaderboard evaluates realism but not diversity. Table~II gives the diversity view: even without an entropy term, Flow-ERD preserves higher CPD than all fine-tuned baselines. This indicates that the gain is not merely a generic fine-tuning effect; ERD reshapes probability mass within the broad support learned by the flow-based AFM backbone. \textcolor{blue}{ERD also surpasses RoaD~\cite{garcia2025road} in both RMM and CPD when both fine-tune the same AFM backbone (\cref{tab:val_cpd}).} Still, $\beta=1.0$ reduces CPD relative to the pretrained AFM backbone with $\Delta \text{CPD} = 0.0174$, consistent with the mode-seeking behavior described in Section~III. Lowering the entropy temperature to $\beta=0.99$ recovers most of the backbone diversity with $\Delta \text{CPD} = 0.003$ while retaining a large realism gain. \textcolor{blue}{Every other fine-tuned method loses at least $0.007$ from its backbone.}

\textcolor{blue}{\Cref{fig:beta_sweep} confirms this trend across the
full sweep: decreasing $\beta$ consistently trades a small amount of
RMM for higher CPD.}
\textcolor{blue}{As a direct validity test of this gain, we recompute \cref{eq:cpd} with the contribution of every vehicle that collides or changes its on-/off-road status during a rollout set to zero (V-CPD); the normalization in \cref{eq:cpd_pair,eq:cpd} is unchanged, so violations can only lower the score. \finalchange{The gate applies only to vehicles; pedestrians and cyclists stay ungated, as they legitimately travel off the road surface (e.g., sidewalks) and in close groups, where these detectors misfire.}
}
\textcolor{blue}{Crucially, most of the diversity gained by lowering
$\beta$ remains after validity gating: from $\beta{=}1.0$ to $0.98$,
CPD increases by $0.0202$ and V-CPD by $0.0189$. For all
$\beta{<}1$, Flow-ERD's V-CPD ($0.167$--$0.180$) exceeds the highest
baseline CPD ($0.1655$, SMART; dashed line). Since V-CPD cannot exceed
CPD, Flow-ERD achieves the highest \finalchange{vehicle-}validity-gated diversity among all
compared methods.}

\textit{3) The preserved diversity reflects intent-level multimodality:}
  \textcolor{blue}{Beyond rollout validity, we assess intent-level diversity by classifying ego-trajectories from 64 rollouts across 1,048 validation scenarios using the WOMD trajectory-type rule~\cite{Ettinger_2021_ICCV}.} 
  In qualitative \Cref{fig:qualitative}(a), SMART assigns almost all of its probability mass to the dominant straight mode. Our $\beta=1.0$ increases diversity among the same intents, and lowering the temperature to $\beta=0.99$ further recovers the rare U-turns.
\Cref{fig:qualitative}(b) \textcolor{blue}{shows that intent entropy
increases monotonically as $\beta$ decreases.}

\textcolor{blue}{\textit{Inference efficiency:}}
\textcolor{blue}{On a single NVIDIA A100 GPU, Flow-ERD generates an $8$\,s closed-loop rollout for all scene agents in $2.09$\,s on average over $256$ validation scenes, i.e., $26.1$\,ms per $0.1$\,s step and $3.83\times$ faster than real time.}

\section{Conclusion}
\label{sec:conclusion}
We presented Flow-ERD, a multi-agent traffic simulator that jointly improves closed-loop realism and rollout diversity. Its AFM backbone generates continuous action sequences while enforcing type-specific kinematic transitions, and ERD fine-tunes the closed-loop rollout distribution with an entropy-regularized distribution-matching objective. Experiments on WOSAC show that Flow-ERD achieves state-of-the-art realism on the test split and dominates the validation realism--diversity Pareto frontier under CPD. These results suggest that diversity should be evaluated explicitly rather than inferred from realism alone. 

\bibliographystyle{\FinalRoot IEEEtran}
\bibliography{\FinalRoot IEEEabrv,\FinalRoot refs}

\end{document}